\pdfoutput=1

\documentclass[11pt]{article}

\usepackage{CJKutf8}

\usepackage{algpseudocode}
\usepackage{algorithm}

\usepackage{fancyvrb}

\usepackage{acl}

\usepackage{times}
\usepackage{latexsym}

\usepackage[T1]{fontenc}
\usepackage[utf8]{inputenc}

\usepackage{microtype}

\usepackage{inconsolata}

\usepackage{bbding}

\usepackage{graphicx}
\usepackage{multirow}
\usepackage{multicol}
\usepackage{booktabs}
\usepackage{amsmath}
\usepackage{caption}
\usepackage[bottom]{footmisc} 
\title{From Text to CQL: Bridging Natural Language and \\Corpus  Search Engine}

\author{Luming Lu$^{1*}$ \
  Jiyuan An$^{1*}$ \
  Yujie Wang$^{2}\thanks{\ \ Equal contribution.}$ \ \ 
  Liner yang$^{1}\thanks{\ \ Corresponding author.}$ \ \  
  Cunliang Kong$^{3}$ \ 
  \textbf{Zhenghao Liu}$^{4}$ \\
  \textbf{Shuo Wang}$^{3}$ \ 
  \textbf{Haozhe Lin}$^{3}$ \ 
  \textbf{Mingwei Fang}$^{1}$\ 
  \textbf{Yaping Huang}$^{2}$ \ 
  \textbf{Erhong Yang}$^{1}$ \\
  $^1$Beijing Language and Culture University, China \quad   
  $^2$Beijing Jiaotong University, China \\
  $^3$Tsinghua University, China \quad 
  $^4$Northeastern University, China \\
  }

\begin{document}
\begin{CJK*}{UTF8}{gbsn}
\maketitle
\begin{abstract}
Natural Language Processing (NLP) technologies have revolutionized the way we interact with information systems, with a significant focus on converting natural language queries into formal query languages such as SQL. However, less emphasis has been placed on the Corpus Query Language (CQL), a critical tool for linguistic research and detailed analysis within text corpora. The manual construction of CQL queries is a complex and time-intensive task that requires a great deal of expertise, which presents a notable challenge for both researchers and practitioners. This paper presents the first text-to-CQL task that aims to automate the translation of natural language into CQL. We present a comprehensive framework for this task, including a specifically curated large-scale dataset and methodologies leveraging large language models (LLMs) for effective text-to-CQL task. In addition, we established advanced evaluation metrics to assess the syntactic and semantic accuracy of the generated queries. We created innovative LLM-based conversion approaches and detailed experiments. The results demonstrate the efficacy of our methods and provide insights into the complexities of text-to-CQL task.
\end{abstract}

\section{Introduction}

Natural Language Processing~(NLP) technologies have significantly improved our interaction with information systems, enabling a more intuitive and effective interface to communicate with computers. Among these advances, the conversion of natural language queries into query languages, such as Structured Query Language~(SQL) for databases, has been a focal point of research. The exploration of linguistic corpora has benefitted significantly from advances in query languages, enabling researchers and practitioners to navigate and analyze text corpora efficiently. While several query languages, such as SQL for databases and various Domain-Specific Languages (DSLs) for other applications, have seen extensive study and application, the focus on Corpus Query Language (CQL) has been relatively less pronounced. Existing research has extensively explored Text-to-SQL\citep{zhong2017seq2sql, liu2022tapex, yu-etal-2018-spider, li2023resdsql, gao2023texttosql, pourreza2023dinsql, dong2023c3, li2023can} and Text-to-DSL\citep{wang2023grammar, staniek2023texttooverpassql} tasks, demonstrating the feasibility and efficiency of translating natural language instructions into formal query statements to interact with databases and information systems.

\begin{figure}[t]
    \centering
    \includegraphics[width=3in]{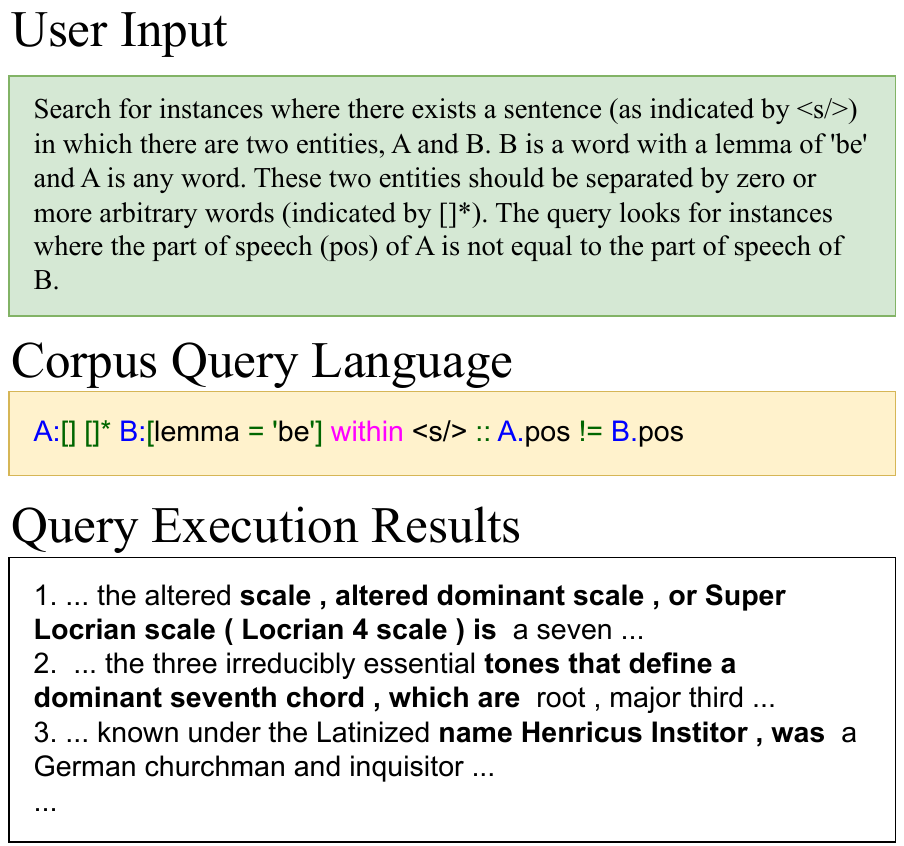}
    \caption{Example of text-to-CQL . Given any input natural language query description, the model is expected to convert it into the corresponding Corpus Query Language (CQL) and the generated CQL should be able to be accurately executed by the Corpus Engine. The CQL uses symbols (in green) with a small number of CQL uses symbols (green) and a small number of keywords (purple) to construct queries, and allows to specify names (blue) for tokens to constrain relationships between tokens. The corpus search engine will return a query execution result.}
    \label{fig:cql}
\end{figure}

CQL is vital for linguistic research as it offers a nuanced approach to querying annotated text corpora, allowing for sophisticated searches based on linguistic features. This capability is crucial for conducting detailed linguistic analysis and supports a wide range of research activities in NLP and computational linguistics. However, crafting CQL queries manually is a time-consuming and error-prone process that requires a high level of expertise in both the query language and the specific annotations of the corpus being queried.

This work introduces the task of text-to-CQL. This task aims to bridge the gap between natural language descriptions and their corresponding CQL representations, facilitating more accessible and efficient interactions with linguistic corpora with rich linguistic annotations. However, unlike its counterparts in text-to-CQL task, the text-to-CQL task faces unique challenges, including a scarcity of dedicated training data and the intricate syntax and semantics of CQL, which are not easily handled by even the most advanced generative models, such as GPT-4, without specialized training and adaptation.
Models need not only to understand the semantics of natural language descriptions, but also to navigate the complexities of linguistic annotations and the specific query constructs of CQL.

This work aims to address these challenges by proposing a comprehensive framework for the text-to-CQL task. We introduce a novel dataset specifically curated for this task, along with methodologies and evaluation metrics tailored to the unique requirements of CQL query generation. Our contributions include the creation of a large-scale dataset that encompasses a wide range of linguistic phenomena and query types, the development of models that adapt large language models and introduce new LLM-based approaches for text-to-CQL task, and the establishment of evaluation metrics that go beyond traditional measures to assess the syntactic validity and semantic correctness of the generated queries.

In summary, our key contributions are as follows:
\begin{itemize}
    \item A large-scale, diverse dataset for text-to-CQL task, providing a benchmark for model evaluation.
    \item A series of LLM-based text-to-CQL methodologies, including both prompt engineering and fine-tuning pretrained language models.
    \item New evaluation metrics designed to accurately reflect the complexities of the text-to-CQL task, focusing on syntactic validity and semantic correctness.
    \item Comprehensive experiments and analysis that highlight the effectiveness of our proposed methods and offer insights into the challenges of text-to-CQL conversion.
\end{itemize}

We will release all our code and datasets for research purposes on Github.

\section{Background}

 Corpus Query Language (CQL) is a query language specifically used to query a corpus with the linguistic features required by users. The CQL utilized in this article is mainly related to the BlackLab~\citep{blacklab}\footnote{https://github.com/INL/BlackLab} corpus Query Language\footnote{ The existing systems for Corpus Query Languages are offshoots of the Corpus Query Language Processor (CQP)~\citep{hardie2012cqpweb} query language, which is a suite of languages designed for the retrieval of lexical information. 
}. 
 
 Numerous corpus search tools endorse and employ CQL, including well-known platforms such as CQPweb\footnote{https://cwb.sourceforge.io/cqpweb.php}, Sketch Engine\footnote{https://www.sketchengine.eu/}, and BlackLab, among others. Some examples of CQL are shown in Table \ref{tab:CQLexample}.

\begin{table*}
    \centering
    \begin{tabular}{p{5em}p{15em}p{15em}} 
    \hline
    \textbf{Type} & \textbf{CQL} &\textbf{NL} \\
    \hline
    Simple & \verb|[lemma="teapot"]| &Find the lemma teapot.\\
    Within & \verb|[pos="N.*"]| \verb|within| \verb|[pos="VB.*"]| \verb|[]|\verb|{0,5} |\verb|[pos="VB.*"]| & Searches for nouns that appear between two verbs to be, the verbs are at a distance of max. 5 tokens from each other.\\
    Condition & \verb|1:[] 2:[] :: 1.pos = 2.pos| & Find any two tokens whose tag is the same.\\
    \hline
    \end{tabular}
    \caption{Example of the Corpus Query Language. The above examples and explanations are all from the Sketch Engine documentation.}
    \label{tab:CQLexample}
\end{table*}

\subsection{CQL Statementes}
\citet{pourreza2023dinsql} categorized SQL into simple, complex, and nesting classes, delineating distinctions in sentence structure complexity within the Text-to-SQL task. Analogously, CQL can be classified into three categories based on keywords, as distinct keywords induce alterations in the CQL structure.

\textbf{Simpe Query.} In CQL, users possess the ability to formulate queries that target the desired corpus by using sequential associations among the tokens. For example, to retrieve instances of research categorized as nouns within the corpus, a user can use the following CQL:
\verb|[word='research' & pos='NN']|

In the above example, we employed a corpus aligned with the Penn Treebank (PTB) \citep{ptb} part-of-speech system. The model's capacity to accurately associate the lexical properties of natural language representations with the appropriate lexical labels represents a potential challenge.

\textbf{Within Query.} As shown in Table \ref{tab:CQLexample}, the "within" syntax serves to partition a CQL statement into two sub-queries, restricting the target retrieved in the initial portion to the scope delineated in the subsequent segment. Typically, "within" is accompanied by a subquery with a larger maximum target length or certain XML structure. 

\textbf{Condition Query.} A condition statement is employed to compare the tokens with each other and to impart additional options to individual tokens. All attributes of a token are eligible for comparison within a condition.

\section{Dataset Construction }

Given a parallel dataset of natural language descriptions and CQL queries $D = \left\{(X_i, Y_i)\right\}_{i=1}^N$, where $X_i = \left\{w_1, w_2, \ldots, w_{n_1}\right\}$ is a query described in natural language and $Y_i = \left\{t_1, t_2, \ldots, t_{n_2}\right\}$ is the CQL corresponding to the NL. $n_1$ represents the length of the natural language description while $n_2$ represents the length of the CQL query. The goal of the text-to-CQL task is to train a model that converts a natural language query description into a query language. This also means that the model needs to have the ability to extract key information from a natural language description and combine it into CQL with the correct syntax. For this purpose, constructing a dataset from natural language to CQL is necessary.

\subsection{Corpus Collection}
We employed two distinct corpora for our study, one in Chinese and the other in English. Both corpora were annotated using Stanford Corenlp.

\textbf{TCFL Textbook.} We collected the main teaching materials to teach Chinese as a foreign language on the market and constructed the TCFL Textbook.

\begin{table}[h!]
\centering
\begin{tabular}{lcc}
\toprule
\textbf{Dataset} & \textbf{Sentences} & \textbf{Tokens}\\
\midrule
TCFL Textbook & 578.4 k & 7.7 M  \\
EnWiki & 138.6 M & 3.1 B  \\
\bottomrule
\end{tabular}
\caption{The token and sentiment size of the corpus we used.}
\label{tab:corpusScale}
\end{table}

\textbf{EnWiki.} We use the EnWiki \citep{enwiki}\footnote{ https://dumps.wikimedia.org/} corpus and clean and extract the text in it using wiki-extractor\footnote{https://github.com/attardi/wikiextractor}\citep{Wikiextractor2015}.  In Table \ref{tab:corpusScale}, We show the size of the cleaned dataset.

Our text-to-CQL dataset is divided into two parts: Chinese NL-CQL pairs based on the TCFL textbooks and English NL-CQL pairs based on Wiki.
\subsection{CQL Generation Strategies}

Certain conventional dataset construction methods, such as the data mining techniques employed in WikiTable\citep{wikitable}, are precluded due to the limited availability of pertinent information on the Internet. Consequently, we develop a novel data collection approach grounded in Chinese collocation extraction.

\subsubsection{Collocation Extraction}

Our approach to data augmentation is based on Chinese Collocation Extraction\citep{hucollocation}. An example of Chinese collocation extraction is shown in 
the Appendix.
The collocation set extraction methodology takes advantage of both surface and dependency relation knowledge, along with statistical methods. Furthermore, we incorporate the enhanced Chinese dependency\citep{yujingsi} to improve the efficacy of collocation extraction. Specifically, our process involves initially employing Stanford CoreNLP\citep{corenlp} for the dependency analysis of sentences within the corpus. Subsequently, enhanced dependencies are introduced, and collocations are extracted from the entire corpus. Finally, a random selection is made from the extracted collocations and applied in classified CQL templates.

\subsubsection{CQL Template}
As depicted in Table \ref{tab:CQLexample}, the CQL queries can be categorized into three distinct types: simple, within, and condition. In alignment with these three types of CQL, we established distinct templates that are tailored for each type.

\textbf{Simple Statements. }
Given that the collocations extracted from the corpus consist of word combinations that exceed two words, our post-extraction procedure involves traversing the word sequence.
Initially, we randomly assign a set of conditions for each token. For any token (such as the noun 'book'), we randomly convert it into the following forms:

1. Simple word query (W).

2. Simple part-of-speech query (P).

3. Query its word and POS at the same time. The logical relationship between the two conditions is randomly chosen from \emph{AND} (WAP) or \emph{OR} (WOP).

4. Query two words at the same time and there is an OR relationship between them (WW). In this case, we also randomly restrict its part of speech with an AND relation (WWP). Another candidate word is selected based on the synonyms specified in the synonym forest.
\begin{table}
    \centering
    \begin{tabular}{p{5em}p{12em}} 
    \hline
    \textbf{Query Type} & \textbf{CQL} \\
    \hline
    W& \verb|[word='book']| \\
    P& \verb|[pos='NN']|\\
    WOP& \verb^[word='book'|pos='NN']^\\
    WAP& \verb^[word='book'&pos='NN']^\\
    WW& \verb^[word='book'|word=^ \verb^'notebook']^\\
    WWP& \verb^[(word='book'|word=^ \verb^'notebook')&pos='NN']^\\

    \hline
    \end{tabular}
    \caption{An example of a CQL Token Queries transformed from Token extracted from a corpus, containing 6 random transformations: simple Word Query (W), Simple Pos Query (P), Word and Pos Query (WAP), Word or Pos Query (WOP), Word or Word query (WW) and Words with Pos Query (WWP)}
    \label{tab:simpleTokenExample}
\end{table}

Examples are shown in Table \ref{tab:simpleTokenExample}. After converting the word in each collocation to CQL, we randomly add empty tokens to it as shown in Algorithm \ref{alg:cap}, where the $mutate$ function refers to the process of converting the collocation word to a CQL token as described in this section, and the $insert\_null\_token$ method randomly adds an unrestricted token to the end of a CQL and assigns it a random number of repetitions or quantifiers using regular expressions.

\begin{algorithm}
\caption{Generation of simple CQL}\label{alg:cap}
\begin{algorithmic}
\State \textbf{Input} $Collocation = \left\{ w_1, w_2, \ldots, w_n \right\}$
\State \textbf{Output} CQL

\State CQL$\gets None $ 
\While{$i \neq 0$}
\If{$freq(w_i) \leq 5 $}
    \State $Abandon()$
\Else
    \State CQL.append(Mutate($w_i$))
    \If{$w_{i+1} = "X"$}
        \State CQL.insert\_null\_token()
        \State $i \gets i + 1$    
    \Else
        \If{$random\_number < 0.5$} 
            \State CQL.insert\_null\_token()
        \EndIf
    \EndIf
    \State $i \gets i + 1$
\EndIf
\EndWhile
\end{algorithmic}
\end{algorithm}

\textbf{Within Statements. }
Two potential subqueries are permissible following the keyword \verb|within|:
 1) \textbf{Simple CQL Subquery.} This causes the corpus searching engine to search for CQL before the \verb|within| keyword within the specified subquery scope. 
 2) \textbf{Structure.} One may utilize XML Structure to confine the query scope to the specified XML domain, with strict prohibition on extending beyond the boundaries defined by the tags. 

For both cases, we randomly apply one of them. On the one hand, two CQLs are generated through collocation analysis, in which the maximum token length they can reference is examined. Then the shorter one is placed preceding the \verb|within| keyword. On the other hand, we randomly specify a certain level of XML format and place it after the \verb|within|. 

We also generate muiti "within" statements. However, nesting multiple levels of queries may lack meaningful interpretation and pose challenges in natural language description. Furthermore, XML structures are commonly segmented at the sentence level and beyond, rendering queries across XML structures practically insignificant. Consequently, we restrict the generation of nested queries to those comprising two "within" keywords, with the XML structure query positioned at the end of the query (representing the highest-priority decision). An example of our generated \emph{within} CQL is shown in the Appendix. 

\textbf{Condition Statements. }
we extract the analytical outcomes of all sentences within the corpus. From these results, we identify token pairs within sentences where parts of speech or words share equality. Subsequently, sentences containing such token pairs are randomly selected, and CQL with condition syntax is generated based on these equivalence relationships. Given that the collocation-based method is no longer applicable to CQL with equivalence relationships, our consideration is limited to scenarios that involve the embedding of CQL within the XML structure in condition statements. An instance of our generated condition CQL is shown in the Appendix.

\subsection{Annotation}

We create the text-to-CQL dataset \textbf{TCQL} with manual annotation. To ensure clarity and precision in natural language descriptions, we implemented a training and selection process for our annotators. Of the initial pool of 14 recruited annotators, we assessed their abilities and ultimately retained eight annotators for subsequent annotation tasks. These annotators have undergraduate degrees and are familiar with both computer science and linguistics.

We perform 4 rounds of labeling for each dataset. First, for the CQLs that have been generated, we perform the initial labeling using the OpenAI GPT-4 API~\citep{openai2023gpt4}. We ask GPT-4 to generate the demand text based on a given CQL and prompt it with the necessary information. Then, we ask the annotator to perform 2 rounds of re-labeling to revise the errors in the results of the initial annotation. Finally, two reviewers who are well-versed in CQL syntax are responsible for reviewing the annotation results again. The size of the labeled data set is shown in Table \ref{tab:combined_statistics}.

\begin{table}[h!]
\centering
\begin{tabular}{lccc}
\toprule
 & \textbf{Train} & \textbf{Dev} & \textbf{Test} \\
\midrule
\textbf{Simple} & 5,631 & 805 & 1,609 \\
\textbf{Within} & 2,332 & 334 & 667 \\
\textbf{Condition} & 1,399 & 199 & 401 \\
\textbf{All} & 9,362 & 1,328 & 2,677 \\
\bottomrule
\end{tabular}
\caption{Combined Classification Statistics for TCQL datasets.}
\label{tab:combined_statistics}
\end{table}

\section{Methodology}
In the construction of the text-to-CQL dataset, we implemented five distinct methodologies, encompassing approaches based on the In Context Learning (ICL) method and approaches using pretraining or fine-tuning pretrained language models.

\subsection{In-Context Learning (ICL) Methods}

We investigate three classifications of Large Language Model (LLM) prompt methods to assess the efficacy of LLMs in text-to-CQL tasks.

\textbf{Documentation Prompt (DP).} Furnish the LLM with a CQL tutorial created by human experts, derived from tutorials accessible in Sketch Engine\citep{sketchengine1,sketchengine2} and Blacklab\citep{blacklab} Documentation. Within the tutorial, we elucidate the syntax of CQL using natural language and furnish illustrative instances, sourced from the tutorial documentation.

\textbf{Few-shot ICL.} We adhere to the methodology outlined by \citet{sun2023battle} and three sets of experiments with different numbers of examples were set up. In each group of examples, we set an example for each of the three types of CQL. 1-Shot Learning (1SL) and 3-Shot Learning (SL) mean that we embed one or three groups of examples in the prompt. Prompt details can be found in the Appendix.

\subsection{Fine-tuning PLM Methods. }  

Models equipped with an encoder-decoder architecture are aptly suited the text-to-CQL task. This category encompasses several models, including BERT\citep{devlin2018bert}， T5\citep{t5}, BART\citep{bart}, and  GPT\citep{radford2019gpt}, among others. this study prioritizes BART due to its integrated encoder-decoder architecture. Furthermore, BART's foundation on a denoising autoencoder pre-training paradigm potentially enhances its proficiency in natural language comprehension and structured query generation, as evidenced by preliminary experimental findings.

For the generation of CQL queries from Chinese texts, this research employed the BART-Chinese model \citep{shao2021bartlargechinese}. We leverage the most expansive `Large' size model available. respectively. Two distinct methodologies were applied for the fine-tuning of the pre-trained language model: Prefix-tuning and Full Model Fine-tuning. The findings indicate that, within the context of the Chinese text-to-CQL task, prefix-tuning yielded superior results. Conversely, for the English text-to-CQL task, full model fine-tuning demonstrated enhanced performance. 
Appendix gives more results of PLM performance on this task. 

\subsection{Metrics}

In this section, we draw upon prior research in the domain of Text-to-SQL, as well as relevant Text-to-Code evaluation metrics, to introduce the four evaluation metrics employed in our study.

\subsubsection{Exact Match (EM)}

Exact Match (EM) is used to evaluate whether the generated SQL query matches exactly the human-annotated standard query. Specifically, the EM metric measures whether the generated SQL query agrees with the reference query without any differences. 
However, execution accuracy may create false positives for CQL queries that are semantically identical but have different forms\citep{yu2018spider, deng-etal-2022-recent}.  

\subsubsection{Valid Accuracy (VA)}

We introduce the Valid Accuracy (VA) metric, which is designed to assess the syntactic correctness of the generated code concerning the CQL grammar. The VA metric provides insights into the model's ability to generate syntactically sound code structures.

\subsubsection{Execution Accuracy (EX)}
Execution Accuracy (EX) metrics are used to evaluate how well the generated CQL query executes on the corpus Engine. It determines whether the generated SQL query executes correctly and returns the desired result \footnote{In our experimental setup, we employ BlackLab as the execution engine for CQL and ascertain the congruence of the corpus retrieval results.}.

\subsubsection{CQLBLEU}

Inspired by~\citet{ren2020codebleu}, we propose new CQLBLEU metrics. This metric is used to assess the similarity between the CQL generated by the model and the reference CQL. Specifically, CQLBLEU is a combination of BLEU\citep{papineni2002bleu} and semantic similarity metrics. Given an candidate CQL $Q_{c}$ and a reference CQL $Q_{r}$, CQLBLEU is defined as:
\begin{equation}
\begin{aligned}
\text{CQLBLEU}(Q_{c},Q_{r}) =& \alpha\cdot \text{BLEU}(Q_{c},Q_{r}) \\ &+ \beta\cdot \text{TS}(Q_{c},Q_{r})
\end{aligned}
\end{equation}
where BLEU stands for BLEU metrics and TS stands for the AST tree similarity which can be a semantic similarity metric. The metric is computed based on the AST generated after syntactic parsing:
\begin{align}
    &T_c = \text{Parse}(Q_{c}) \\
    &T_r = \text{Parse}(Q_r) \\
    &\text{TS}(Q_{c},Q_{r}) = \text{Sim}(T_c,T_r) 
\end{align}
where $T_c$ and $T_r$ are the CQL AST of $Q_{c}$ and $Q_r$ parsed by Blacklab. The Sim function compares each node in the AST of $Q_{c}$ by itself and its direct children for the presence of $Q_r$:
\begin{align}
    \text{Sim}(T_c, T_r)=     \frac{\sum_{n_c \in \text{N}(T_c)} \text{Match}(n_c, T_r)}{|\text{N}(T_c)|}
\end{align}
where $\text{N}(T)$ represents the set of non-leaf nodes in tree $T$, and $\text{Match}(n_c, T_r)$ is a function that returns 1 if a node $n_c$ from $T_c$ and its direct children have a matching signature in $T_r$, and 0 otherwise. The matching criterion for a node $n_c$ with signature $s(n_i) = (\text{N}(n_c), \text{C}(n_c), \text{K}(n_c))$ against $T_r$ is defined as follows:
\begin{equation}
\begin{aligned}
    & \text{Match}(n_i, T_r) = \\ &\begin{cases}
    1, & \text{if } \exists n_r \in \text{N}(T_r) : s(n_i) = s(n_r) \\
    0, & \text{otherwise}
    \end{cases}
\end{aligned}
\end{equation}

The coefficients $\alpha$ and $\beta$ in the definition of CQLBLEU allow for balancing the contribution of syntactic similarity, as measured by BLEU, and semantic similarity, as measured by $TS$, to the overall metric. In our work, we choose $\alpha = 0.5$ and $\beta = 0.5$. 

\section{Analysis}

\begin{table*}[ht]
\centering
\begin{tabular}{lp{2em}cccccccc}
\toprule
\multirow{2}{*}{Model} & \multirow{2}{*}{Settings} & \multicolumn{4}{c}{TCFL Textbook} & \multicolumn{4}{c}{EnWiki} \\
\cmidrule(lr){3-6} \cmidrule(lr){7-10}
      &          & EM & VA & EX & CQLBLEU & EM & VA & EX & CQLBLEU \\
\midrule
BART-Chinese & - & 46.52 & 80.46& 50.95 & 72.95 & - & -& -& - \\
BART-English         & - & - & - & - & - & 37.58 & 81.74 & 44.30 & 82.13 \\
GPT-4 & DP & 35.17 & 77.52 & 51.79 & 74.95 & 14.93 & 75.37 & 24.49 & 67.63 \\
GPT-4 & 1SL & 47.81 & 81.84 & 62.71 & 82.22 & 43.31 & 82.24 & 51.87 & 82.93 \\
GPT-4 & 3SL & 67.49 & 90.28 & 77.85 & 91.83 & 58.24 & 89.74 & 65.53 & 89.93\\
\bottomrule
\end{tabular}
\caption{Experiment results. The table contains the results of four evaluation metrics: Exact Match (\textbf{EM}), Valid Accuracy (\textbf{VA}), Execution Accuracy(\textbf{EX}), and \textbf{CQLBLEU}. We choose the BART-Large model and use its Chinese branch to fit our Chinese dataset.}
\label{tab:model_performance}
\end{table*}

\subsection{LLM capability assessment}
In our LLM-based ICL experiments, we found three significant features of LLM for this task:

\textbf{LLM by itself is almost incapable of writing CQL correctly}. In our early test, LLM shows low performance of several methods for each of the three CQL classifications: simple, within, and conditional. LLM did not perform better than PLM even when Documentation Prompt (DP) was provided. This may be due to the fact that the training data that LLM was exposed to may have contained fewer CQL examples, and these examples were mostly focused on simple classification, and not much on the other two classifications, which are more flexible and broader in application scenarios.

\textbf{LLM is much better at learning from examples.} We experimented with having LLM learn CQL knowledge from documents written by human experts (DP) and having LLM learn from examples given CQL-NL pairs (1SL and 3SL). In the DP approach, there is still a large gap between LLM and finetuned PLM, which may mean that LLM is not as efficient at reading documents that are more easily understood by humans. The results show that CQL can effectively understand the syntax of the query language in fewer samples. This is consistent with \citet{staniek2023texttooverpassql}'s conclusion.

\textbf{LLM understands the semantics expressed in human language}. In most cases, LLMs achieve high CQLBLEU scores even if they are not given detailed hints about the CQL syntax or if their execution results do not meet expectations. This means that LLM writes answers that are closer to human answers in terms of semantic similarity and text. This ability of LLM can continue to be enhanced with more hints or examples. This also confirms that LLM learns not only formal knowledge from examples but also semantic information.

\subsection{PLM Performance Analysis}

\subsubsection{Performance of PLM on different languages}
Based on the experimental results described in the previous section, the performance of the same large-sized BART model shows differences in the text-to-CQL tasks for both English and Chinese languages. Beyond the differences in model performance due to the language used for fine-tuning, we believe a more significant reason is the addition of the "lemma" attribute in English CQL compared to Chinese. In addition to "word" and "pos" in Chinese queries, English queries also include "lemma," requiring the model to learn an additional attribute name. Furthermore, the forms of words and their lemmas are quite similar in natural language expression and are often mixed in actual human queries. The model exhibits similar behavior, where the predicted CQL queries differ from the gold standard only in the attribute names "word" and "lemma," which is a very common type of error occurrence.

\subsubsection{Performance of PLM on different query difficulties}

To better assess the performance of our proposed model, we categorized CQL queries into three levels of difficulty based on human habits in writing CQL queries. According to our intuition, the difficulty of generating CQL queries from text for the model should follow the order: Simple < Within < Condition. However, the model's performance in some cases deviated from our expectations, showing a significantly better performance on \emph{condition} type than on \emph{within} type (for example, when using the DP method). 
To elucidate the reasons behind this phenomenon, we provide detailed statistical data from the dataset, as shown in the Appendix.
We found that in terms of the character length of CQL queries and the number of constraint conditions, \emph{Within} type far exceeds Condition type, implying that natural language inputs of the Within type lead to the generation of the longest CQL queries with the most constraint conditions, which typically signifies a higher probability of errors. Conversely, \emph{condition} type demonstrated more complex query logic, but since it involves more non-constraint word queries, the primary challenge it poses to the model is the understanding of the logic in the natural language input rather than longer CQL queries and more constraint conditions.

\section{Related Work}
\label{sec:related_work}

\subsection{Text-to-SQL}
\label{sec:text2sql}

Text-to-SQL task, a key research area, involves translating natural language questions into SQL queries. 
Seq2SQL~\citep{zhong2017seq2sql} is a notable model in this field, utilizing policy-based reinforcement learning to accurately generate SQL queries, particularly focusing on the unordered nature of query conditions. 
It excelled in both execution and logical form accuracy on the WikiSQL dataset.
In the same data set, TAPEX~\citep{liu2022tapex}, an execution-centric table pretraining approach that learns a neural SQL executor over a synthetic corpus, achieved the state-of-the-art results.

The Spider~\citep{yu-etal-2018-spider} dataset furthered Text-to-SQL research by presenting a complex, cross-domain semantic parsing challenge. 
It features varied SQL queries and databases, pushing models to adapt to new structures and databases. 
RESDSQL~\citep{li2023resdsql} introduced a ranking-enhanced encoding and skeleton-aware decoding framework that effectively decouples schema linking and skeleton parsing, demonstrating improved parsing performance and robustness on the Spider dataset and its variants.

Recent advances in large language models (LLMs), such as GPT-4~\citep{openai2023gpt4} and Claude-2, have also shown impressive results in this domain~\citep{gao2023texttosql, pourreza2023dinsql, dong2023c3}. 
To our knowledge, most previous benchmarks, including Spider and WikiSQL, focused on database schemas with limited rows, creating a gap between academic studies and real-world applications. 
To bridge this gap, the BIRD~\citep{li2023can} benchmark was introduced, providing a comprehensive text-to-SQL dataset that emphasizes the challenges of dealing with dirty and noisy database values, grounding external knowledge, and ensuring SQL efficiency in massive databases.

However, adapting these methods from Text-to-SQL to text-to-CQL isn't straightforward, primarily because of the scarcity of training data for text-to-CQL. 
This challenge motivated the proposal of this paper.

\subsection{Text-to-DSL}
\label{sec:text2dsl}

The field of generating Domain-Specific Languages (DSLs) from natural language, known as Text-to-DSL, has seen a surge in interest, primarily due to the emergence of LLMs capable of understanding and generating structured languages. 
A notable approach in this area is Grammar Prompting~\citep{wang2023grammar}, which leverages Backus–Naur Form (BNF) grammars to provide LLMs with domain-specific constraints and external knowledge. 
This method has shown promise across various DSL generation tasks, including semantic parsing and molecule generation.

Text-to-OverpassQL~\citep{staniek2023texttooverpassql} focused on generating Overpass queries from natural language. 
This task is particularly challenging due to the complex and open-vocabulary nature of the Overpass Query Language (OverpassQL). 
\citet{staniek2023texttooverpassql} proposed the OverpassNL dataset and established task-specific evaluation metrics.

\section{Conclusion}

In this paper, we introduce a novel task, text-to-CQL, aimed at converting natural language input into Corpus Query Language (CQL). This task holds significant relevance for corpus development and research, sharing certain similarities with existing text-to-query language tasks. The text-to-CQL task, however, presents distinctive challenges owing to its unique syntax and limited availability of publicly accessible resources. To support research in this domain, we propose TCQL—a template-based generation approach for creating text-to-CQL datasets. To ensure the authenticity of the dataset, we build the data based on NLP tasks such as collocation extraction and lexical labeling. We use this dataset to test the performance of several state-of-the-art models, propose a new evaluation metric, CQLBLEU, based on N-gram similarity and AST similarity, and build a baseline for the tasks with reference to several commonly used metrics in Text-to-SQL.We evaluate the results in detail, revealing the strengths and weaknesses of the considered learning strategies. We hope that this contribution will positively impact corpus development and applications, advancing technology in both the realms of NLP and linguistics.

\label{sec:bibtex}

\section*{Limitations}
This work introduces a novel approach to converting natural language queries into Corpus Query Language (CQL) expressions. Despite its potential to significantly advance research in corpus linguistics and natural language processing, several limitations must be acknowledged:

\begin{itemize}
    \item Currently, the construction of the TCQL dataset used in this paper is based on automatically generated and manually labeled due to the lack of a large amount of raw CQL data generated from real human queries. Despite the fact that we have used a variety of methods to enhance its authenticity, it is still possible to generate queries that are not meaningful enough.
    \item The scalability of the proposed solution to longer text queries and its dependency on computational resources are concerns that may limit its applicability in resource-constrained settings.
\end{itemize}

Future research is encouraged to address these limitations, exploring the method's applicability to a wider range of languages, enhancing its scalability, and reducing its computational requirements.

\bibliography{custom}

\end{CJK*}
\end{document}